%% file: main.tex
\documentclass[10pt,twocolumn,letterpaper]{article}

\usepackage{wacv}              

\usepackage{multirow}   
\usepackage{amssymb}    
\usepackage[table]{xcolor}
\usepackage{colortbl}
\usepackage{amsmath,amssymb,amsthm}
\usepackage{mathtools}
\usepackage{bm}

\usepackage{booktabs}
\usepackage{enumitem}
\usepackage{xcolor}
\usepackage{wrapfig}
\newcommand{\R}{\mathbb{R}}

\definecolor{ours}{RGB}{230,245,255}
\definecolor{gray}{RGB}{220,220,220}


\definecolor{wacvblue}{rgb}{0.21,0.49,0.74}
\usepackage[pagebackref,breaklinks,colorlinks,allcolors=wacvblue]{hyperref}

\def\wacvPaperID{*****} 
\def\confName{WACV}
\def\confYear{2027}

\title{Query-Conditioned Evidential Keyframe Sampling\\ for MLLM-Based Long-Form Video Understanding}

\author{
Yiheng Wang\textsuperscript{1}\thanks{Equal contribution.},
Lichen Zhu\textsuperscript{1}\footnotemark[1],
Yueqian Lin\textsuperscript{1},
Yudong Liu\textsuperscript{1},\\
Jingyang Zhang\textsuperscript{2},
Hai ``Helen'' Li\textsuperscript{1},
Yiran Chen\textsuperscript{1}\\
\textsuperscript{1}Duke University, Durham, NC, USA\\
\textsuperscript{2}Independent Researcher\\
{\tt\small \{yiheng.wang, lichen.zhu, y.lin, yudong.liu, hai.li, yiran.chen\}@duke.edu}
}

\begin{document}
\maketitle
\input{sec/0_abstract}    
\input{sec/1_intro}
\input{sec/2_related_works}
\input{sec/3_methodology}

\input{sec/4_experiments}
\input{sec/5_conclusion}

\section*{Acknowledgments}
This work was supported in part by NSF 2625580, NSF IIS-2332744, and ARO W911NF-23-2-0224.
We also thank the Gemini Academic Program for providing API credits through the Gemini Academic Program Award.

{
    \small
    \bibliographystyle{ieeenat_fullname}
    \bibliography{main}
}

\end{document}

%% file: sec/0_abstract.tex
\begin{abstract}
Multimodal Large Language Models (MLLMs) have shown strong performance on video question answering, but their application to long-form videos is constrained by limited context length and computational cost, making keyframe sampling essential. Existing approaches typically rely on semantic relevance or reinforcement learning, which either fail to capture evidential clues or suffer from inefficient combinatorial optimization.
In this work, we propose an evidence-driven keyframe sampling framework motivated by an information-theoretic perspective.
We consider maximizing the conditional mutual information between the selected frames and the answer given the query as an idealized objective, and use a modular singleton-gain surrogate to motivate a tractable frame-wise selection criterion. Guided by this motivation, we train a query-conditioned evidence scoring network using annotated answer-supporting segments.
A contrastive objective encourages greater aggregate score mass on frames from evidence regions than on background frames, providing a practical evidence-based criterion beyond semantic similarity.
Experiments on long-form video understanding benchmarks show that our method consistently outperforms prior sampling strategies under comparable frame budgets. With Qwen2.5-VL-7B under the same 32-frame budget, our method improves over uniform sampling by 10.1 points on LVBench and 4.4 points on the long-video subset of VideoMME.
\end{abstract}

%% file: sec/1_intro.tex
\section{Introduction}
\label{sec:intro}

Multi-modal Large Language Models (MLLMs)~\citep{slowfast-llava,qwen2vl,qwen3vl,internvl,internvl2,internvl2.5,smolvlm} have demonstrated strong capabilities in video understanding and responding to natural language questions. However, practical deployment remains constrained by limited context length and token budgets. 
Even at a modest sampling rate of one frame per second, a one-hour video contains 3,600 frames, amounting to nearly one million visual tokens when each frame is encoded into 256 tokens. 
This scale far exceeds the practical context window of most multimodal large language models (MLLMs), making exhaustive frame processing infeasible and keyframe sampling essential for long-form video understanding.
This raises a central question for keyframe sampling: which frames truly matter for answering a given query. 

\begin{figure}[t]
    \centering
\includegraphics[width=0.8
\linewidth]{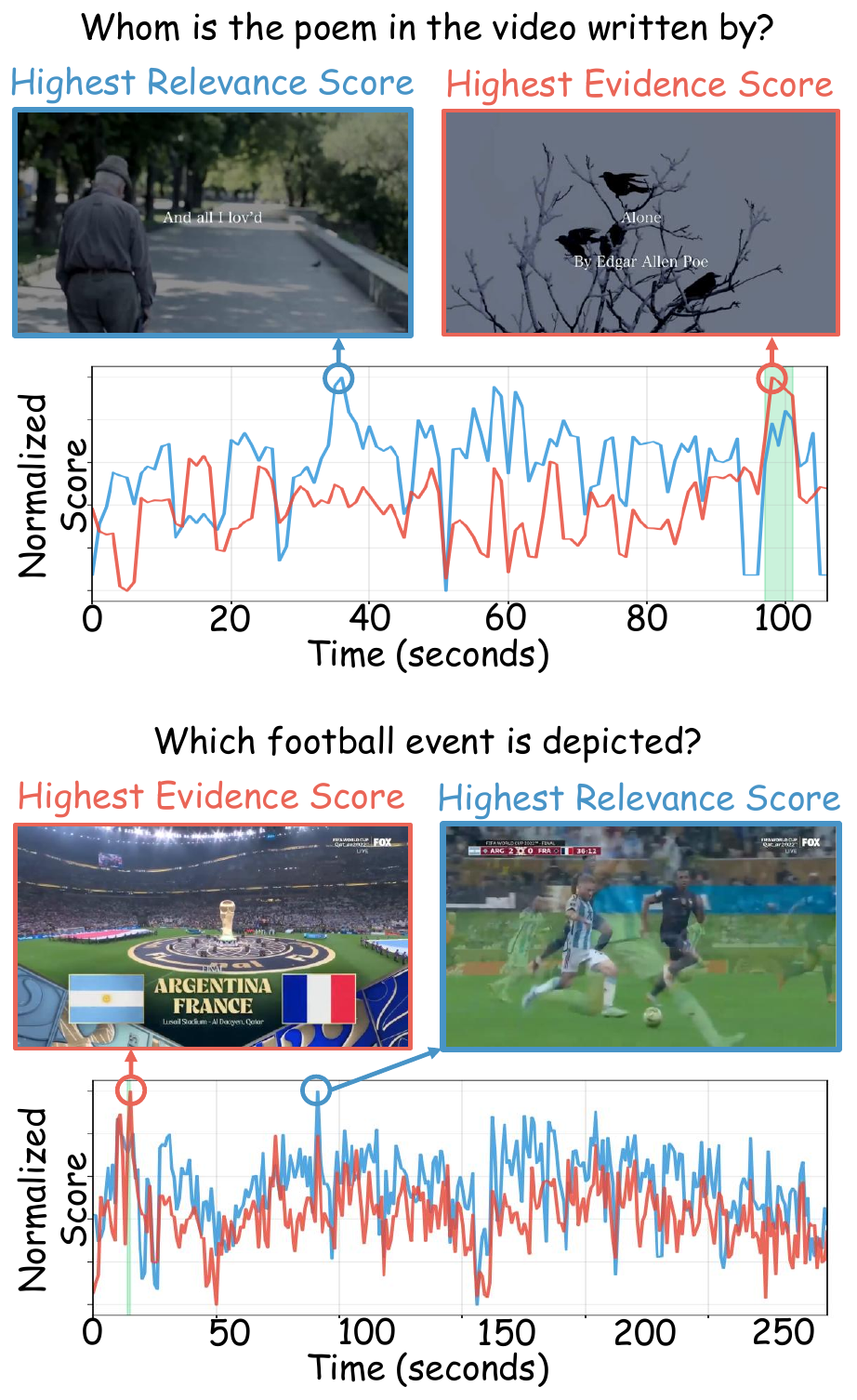}
    \vspace{-10pt}
    \caption{Semantic relevance vs. evidential score in keyframe selection. Scores are min–max normalized.
Frames with the highest semantic similarity (\textcolor{blue}{blue}) predicted by CLIP~\citep{clip}, describing poem content (top) or general match scenes (bottom), are not always informative for answering the query. In contrast, frames with the highest evidence scores (\textcolor{red}{red}) predicted by our method better capture answer-critical information. The \textcolor{green}{green} regions indicate the temporal segments containing the key evidence needed for answering.}
    \label{fig:example}
    \vspace{-10pt}
\end{figure}

A natural approach is to select keyframes based on visual or semantic relevance. However, such relevance does not necessarily imply that a frame provides the evidence required to answer the query~\citep{reason}, as illustrated in Figure~\ref{fig:example}. 
Early keyframe sampling approaches follow this intuition and rely on semantic-matching heuristics~\citep{aks,qframe,focus}, which utilize pretrained vision-language models~\citep{clip,blip} to obtain the similarity score between frames and queries, and sample keyframes with heuristic rules to prioritize visual relevance and temporal coverage. 
However, such strategies often lead to suboptimal performance due to the mismatch between semantic similarity and actual evidential usefulness.
Reinforcement learning (RL)-based keyframe sampling policies~\citep{tspo,reason} mitigate this issue by training policy networks to select keyframes that enable MLLMs to answer queries correctly. While effective, these methods explore an exponential combinatorial search space, leading to poor sample efficiency. Furthermore, the sparse rewards derived from answer accuracy introduce a daunting credit assignment problem, making it difficult to provide precise quality estimations for individual frames, which results in training instability.

To overcome these limitations, we propose an evidence scoring framework from an information-theoretic perspective. Unlike plain CLIP similarity, which measures semantic alignment between an isolated frame and the query, our approach targets answer-supporting visual evidence. We formulate keyframe sampling as maximizing the conditional mutual information between the selected frames and the answer given the query. Under a conditional-independence assumption, the objective is monotone submodular, motivating a modular singleton-gain surrogate that provides a tractable frame-wise selection criterion.

Since the singleton information gain is not directly observable, we instantiate this criterion with a query-conditioned evidence scoring network trained on annotated answer-supporting regions. A contrastive objective encourages greater aggregate evidence-score mass on frames within these regions than on background frames. This region-level supervision avoids sparse answer-level rewards and MLLM-in-the-loop training while enabling evidence-based frame selection beyond semantic similarity.


Our main contributions are as follows.
First, extensive experiments demonstrate consistent improvements under matched frame budgets. With Qwen2.5-VL-7B, our method improves accuracy by 10.1 points on LVBench~\citep{lvbench} and by 4.4 points on the long-video subset of VideoMME~\citep{videomme} under the same 32-frame budget.
Second, we develop a tractable frame-wise selection framework motivated
by an idealized conditional mutual information objective and a modular
singleton-gain surrogate.
Third, we design a query-conditioned evidence scoring network trained with region-level supervision from annotated evidence segments, providing a practical evidence-based criterion beyond semantic similarity.


%% file: sec/2_related_works.tex
\section{Related Works}
\subsection{MLLMs for Video Understanding}
The rapid advancement of MLLMs has significantly pushed the boundaries of video understanding.
VideoLLaMA~\citep{videollama} pioneer the field with vision-language and audio-language pretraining, enabling video-grounded conversation. 
VideoLLaMA2\citep{videollama2} improve the previous work with a spatial-temporal convolution connector.
VideoLLaMA3 \citep{videollama3} further propose a video-centric fine-tuning stage to boost the video understanding performance.
Qwen-Series~\citep{qwen2vl,qwen25vl,qwen3vl} propose M-RoPE to integrate temporal dependency into the attention mechanism, boosting their performance for video understanding and temporal perception. 
InternVL-Series~\citep{internvl,internvl2,internvl2.5} scale up the vision encoder and extend their multimodal framework to video understanding to improve video understanding performance.
Despite these architectural leaps, efficient long-form video understanding remains a critical bottleneck. As video duration extends into hours, the linear accumulation of dense visual tokens inevitably overwhelms context windows, leading to severe computational latency and memory consumption. Consequently, dynamically isolating crucial evidence information from highly redundant temporal streams remains a fundamental open challenge.

\subsection{Keyframe Sampling}
Recently, keyframe sampling methods have emerged to improve MLLMs in long-form video understanding by alleviating the computational burden of processing lengthy visual sequences. These approaches generally fall into two categories: training-free semantic-matching methods and trainable policies. Training-free approaches dynamically select frames based on heuristic rules. Specifically, AKS and FOCUS~\citep{aks,focus} balance semantic relevance and temporal coverage via a plug-and-play optimization algorithm. Q-Frame~\citep{qframe} performs query-aware sampling with dynamic multi-resolution scaling, while MDP3~\citep{mdp3} casts keyframe selection as a training-free list-wise optimization that jointly models frame relevance and inter-frame redundancy. 
These methods provide useful query-aware selection signals. Our work investigates whether supervision from temporal evidence regions can further improve frame ranking beyond pretrained semantic alignment.
Trainable methods optimize frame selection by introducing learnable components or leveraging reinforcement learning (RL). 
Closely related work on visually grounded VideoQA explicitly associates predicted answers with supporting temporal evidence. NG+~\citep{nextgqa} learns to localize answer-supporting moments through Gaussian mask optimization and cross-modal learning. 
GCG~\cite{gcg} derives CLIP-based pseudo-labels from question--answer descriptions and learns Gaussian temporal masks through contrastive grounding.
SeViLA~\cite{sevila} adapts BLIP-2 into a Localizer and an Answerer, using answerer-generated keyframe pseudo-labels to refine localization through self-chaining.
VideoITG~\cite{videoitg} trains a Video-LLM-based selector using instruction-conditioned temporal-grounding annotations generated by an automated pipeline.
Frame-Voyager~\citep{framevoyager} learns to query informative frame combinations ranked by a pretrained Video-LLM, and GenS~\citep{gens} reframes sampling as a generative task that identifies question-relevant frames. In parallel, MLLM-Selector and FFS~\citep{mllm-selector,ffs} incorporate learnable filtering modules and register tokens to adaptively determine frame importance conditioned on the query, while HFS~\citep{hfs} pursues holistic, set-level selection through a differentiable relevance–coverage–redundancy objective. Exploring RL-based paradigms, TSPO~\citep{tspo} trains an event-aware agent via GRPO, while ReaSon~\citep{reason} adopts a causal information bottleneck with counterfactual interventions to identify decisive moments. While effective, these approaches operate over a large combinatorial space, leading to low sample efficiency.

%% file: sec/3_methodology.tex
\section{Methodology}
\subsection{Information-Theoretic Motivation}
\begin{figure*}[t]
    \centering
\includegraphics[width=\linewidth]{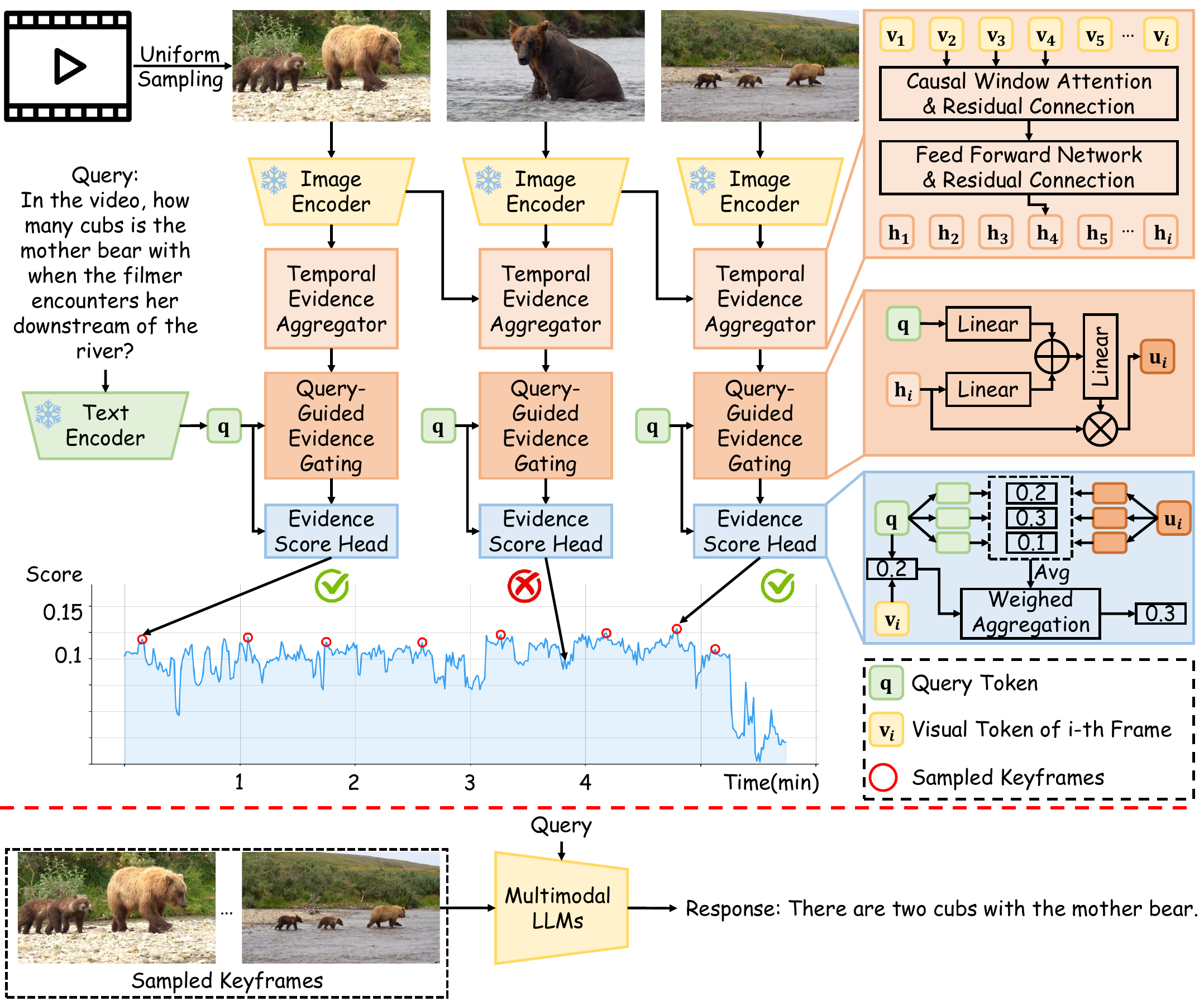}
    \vspace{-20pt}
    \caption{Overview of the query-conditioned evidence scoring network. The input video is uniformly sampled into frames, which are encoded into frame embeddings and scored to obtain frame-level evidence scores conditioned on the query. Frames with the highest evidence scores are then selected and fed into an MLLM to generate the final response.}
    \label{fig:model}
    \vspace{-10pt}
\end{figure*}
\textbf{Problem Statement.}
Given a video represented as a sequence of frames $V = \{f_1, \ldots, f_n\}$ and a natural language query $Q$, keyframe sampling seeks a compact subset $S \subseteq V$ under a frame budget $|S| \leq m$ such that multi-modal large language models can answer $Q$ from $S$ with accuracy comparable to using $V$. 
In the following information-theoretic motivation, frame observations and answers are treated as random variables over the data distribution, and $S$ denotes a collection of selected frame observations. 

As an idealized information-preservation criterion, we consider maximizing the conditional mutual information between the selected
observations and the ground-truth answer, given the query:

%
\begin{equation}
    \max_{S} \; I(S;\, O \mid Q) \quad \text{s.t.} \quad |S| \leq m,
    \label{eq:cib}
\end{equation}
where $O$ denotes the ground truth answer, $m$ denotes the frame budget and $I(S;\, O \mid Q)$ denotes the conditional mutual information. 
The conditioning on $Q$ reflects the fact that the evidential value of a frame is inherently query-dependent. The same frame may be decisive for one question yet irrelevant to another.

To interpret this objective, we expand the conditional mutual
information:
\begin{equation}
I(S;O\mid Q)
=
\mathbb E_Q
\left[
\mathbb E_{p(S,O\mid Q)}
\left[
\log \frac{p(O\mid S,Q)}{p(O\mid Q)}
\right]
\right].
\end{equation}
For a fixed data distribution, $H(O\mid Q)$ is independent of the
selection. Maximizing $I(S;O\mid Q)$ is therefore equivalent to
maximizing the expected log-likelihood under the true conditional
distribution:
\begin{equation}
\max_S \mathbb E[\log p(O\mid S,Q)].
\end{equation}
Here, $p$ denotes the true conditional distribution, rather than the
predictive distribution of the deployed MLLM. This identity motivates
selecting observations that make the answer more predictable; it
does not establish an equivalence with the classification accuracy
of a frozen MLLM. We assess the utility of the learned selection
scores empirically through downstream QA accuracy.
\noindent\textbf{Singleton-Based Scoring Motivation.}
Under the simplifying assumption that candidate-frame observations are conditionally independent given the ground-truth answer and query, $F(S)=I(S;O\mid Q)$ is normalized, monotone, and submodular, following standard information-gain results~\cite{krause2005nearoptimal}.
Consequently,
\begin{equation}
F(S)
\leq
\sum_{f_i\in S}F(\{f_i\})
=
\sum_{f_i\in S}I(f_i;O\mid Q).
\end{equation}
The additive upper bound motivates a computationally tractable frame-wise scoring approach. However, it does not establish equivalence to the original set objective or a near-optimality guarantee for the implemented selector. Moreover, conditional independence need not hold for temporally correlated video frames, and additive scoring does not explicitly account for redundancy or complementary evidence. We therefore use this formulation as motivation for the scoring design.


\noindent\textbf{Learning Practical Evidence Scores.}
\label{sec:mi_to_scoring}
The information-theoretic perspective motivates prioritizing
answer-relevant observations. We operationalize this goal by training
a query-conditioned scorer $g_\theta(f_i,Q)$ with temporal
evidence-region supervision. The scorer learns to distinguish frames
within annotated evidence regions from sampled background frames.
Its output is a supervised ranking signal, not an estimate of
$I(f_i;O\mid Q)$. Although one score is assigned to each frame,
the network may use local temporal context when computing that score.
The architecture and training objective are detailed in \S\ref{sec:architecture} and \S\ref{sec:training}.

\noindent\textbf{Keyframe Selection Rules.}
To promote temporal coverage, we partition the video into \(B\)
equal-length bins \(\{\mathcal{B}_b\}_{b=1}^{B}\). Let
\(\operatorname{Top}_r(\mathcal{B}_b)\) denote the \(r\) frames in
\(\mathcal{B}_b\) with the largest evidence scores.
We select
\begin{equation}
S^\star
=
\bigcup_{b=1}^{B}
\operatorname{Top}_r(\mathcal{B}_b),
\qquad m=Br.
\label{eq:stratified_selection}
\end{equation}
This formulation provides a unified coverage--concentration
trade-off. Setting \(B=m\) and \(r=1\) selects one frame per bin for
temporal coverage, whereas \(B=1\) and \(r=m\) recovers global
top-\(m\) selection.

\subsection{Query-Conditioned Evidence Scoring Network}
\label{sec:architecture}
As shown in Figure~\ref{fig:model}, the scoring network $g_\theta$ comprises a pretrained vision-language encoder, a temporal evidence aggregator, a query-guided evidence gating module, and an evidence score head.

Each frame $f_i$ and the query $Q$ are encoded by a shared vision-language encoder:
\begin{equation}
    \mathbf{v}_i = \mathcal{E}_v(f_i) \in \R^d, \quad \mathbf{q} = \mathcal{E}_t(Q) \in \R^d.
\end{equation}

The scoring network assigns a scalar evidence score to each candidate frame. Computing this score from an isolated frame may overlook useful
local temporal context.
The evidential value of a frame in video is often inseparable from its local temporal context, \textit{e.g.}, whether a person is picking up or putting down an object can only be disambiguated by observing the immediately preceding motion.
Encoding each frame in isolation would thus yield a poor estimate of evidence.

To address this, we introduce a temporal evidence aggregator with a causal window attention mechanism. For frame $f_\tau$ at position $\tau$, we define the causal window
$\mathcal{W}_\tau = \{f_j \mid \max(1,\,\tau-w+1) \leq j \leq \tau\}$.
The contextualized representation is:
\begin{equation}
    \mathbf{h}_\tau
    =
    \mathcal{T}\!\left(
        \mathbf{v}_\tau
        \mid
        \{\mathbf{v}_j : f_j \in \mathcal{W}_\tau\}
    \right).
    \label{eq:causal_window}
\end{equation}
Crucially, the window size $w$ is kept small relative to the video length, confining temporal aggregation to a local neighborhood.

To ground the contextualized visual representation in the query semantics, we introduce a query-guided evidence gating module that acts as a channel-wise information valve:
\begin{equation}
    \mathbf{g}_i = \sigma(\mathbf{W}_h \mathbf{h}_i + \mathbf{W}_q \mathbf{q} + \mathbf{b}), \quad
    \mathbf{u}_i = \mathbf{h}_i \odot \mathbf{g}_i,
\end{equation}
where $\mathbf{W}_h, \mathbf{W}_q \in \R^{d \times d}$, $\mathbf{b} \in \R^d$, and $\sigma$ is the sigmoid function. The gate $\mathbf{g}_i \in (0,1)^d$ suppresses query-irrelevant channels while preserving visual features that are informative for answering the query.

We decompose the evidence score across $H$ semantic subspaces to capture distinct aspects of query--frame alignment. In the $a$-th subspace:
\begin{equation}
    s_{i,a}
    =
    \frac{1}{\gamma_a}
    \frac{
        (\mathbf{W}_v^{(a)}\mathbf{u}_i)^\top
        (\mathbf{W}_q^{(a)}\mathbf{q})
    }{
        \|\mathbf{W}_v^{(a)}\mathbf{u}_i\|_2
        \;\|\mathbf{W}_q^{(a)}\mathbf{q}\|_2
    },
\end{equation}
where $\mathbf{W}_v^{(a)}, \mathbf{W}_q^{(a)}
\in \R^{d_a \times d}$, $d_a=d/H$, and $\gamma_a$ is a learnable temperature. The $\ell_2$ normalization makes scoring invariant to feature magnitudes, relying purely on directional alignment. The overall evidence score aggregates across subspaces:
\begin{equation}
    g_\theta(f_i,Q)
    =
    \lambda
    \frac{\mathbf{v}_i^\top\mathbf{q}}
    {\|\mathbf{v}_i\|_2\;\|\mathbf{q}\|_2}
    +
    (1-\lambda)\frac{1}{H}
    \sum_{a=1}^{H}s_{i,a},
\end{equation}
where $\lambda\in[0,1]$ is a learnable parameter that balances the contribution of the two terms. Incorporating the pretrained vision-language cosine similarity provides a strong semantic-alignment prior for evidence scoring.

\subsection{Training Objectives}
\label{sec:training}
Following the information-preservation motivation, we learn frame-wise evidence scores using temporal evidence-region supervision. The training objective encourages greater aggregate softmax probability mass for frames within annotated evidence regions than for sampled background frames. 

For each query $Q$, we construct a set of positive frames $\mathcal{F}^{+}$ and a set of negative frames $\mathcal{F}^{-}$. 
A frame is labeled positive if it falls within an annotated evidence segment and negative otherwise. The annotations identify answer-supporting temporal regions and therefore provide region-level supervision for distinguishing evidence-containing portions of the video from background.

We train $g_{\theta}$ with a multi-positive contrastive loss~\citep{oord2018representation}. Given the positive set $\mathcal{F}^{+}$ and negative set $\mathcal{F}^{-}$ for a query $Q$:
\begin{equation}
\mathcal{L}
=
-\log
\frac{
\sum_{x \in \mathcal{F}^{+}}
\exp g_{\theta}(x,Q)
}{
\sum_{x \in \mathcal{F}^{+}}
\exp g_{\theta}(x,Q)
+
\sum_{x \in \mathcal{F}^{-}}
\exp g_{\theta}(x,Q)
}.
\end{equation}

By encouraging greater aggregate softmax probability mass on frames from annotated evidence regions than on background frames, this objective promotes the selection of answer-supporting regions while suppressing background content.

%% file: sec/4_experiments.tex
\section{Experiments}
\subsection{Experimental Setup}
\subsubsection{Implementation Details}
We train the proposed keyframe sampling model on the LLaVA-Video subset of the Seeker-173K dataset~\citep{videoo3}. 
We use the temporal evidence-region annotations provided by Seeker-173K, which do not overlap with benchmarks.
The frames sampled within the evidence segments are treated as positive samples, while those outside the segments are considered negative samples.
Our model adopts CLIP-ViT-L~\citep{clip} as the default vision-language encoder. The encoder is kept frozen during training, while the remaining modules, comprising approximately 10M parameters, are optimized. Training is conducted for 5 epochs on 8 NVIDIA RTX A5000 GPUs with a batch size of 128.
We adopt a learning rate schedule with linear warmup over the first 10\% of training steps to a peak of $5 \times 10^{-5}$, followed by cosine decay.
At inference time, candidate frames are sampled at 1 frames per second. After selection, the frames are arranged in chronological order before being passed to the downstream MLLM.
Evaluation is conducted on NVIDIA L40S GPUs.

We evaluate the model on long-video understanding benchmarks, \textit{i.e.}, LVBench~\citep{lvbench} and Video-MME~\citep{videomme}. 
For each video-question pair, our keyframe sampling method is applied to sample $m$ frames, which are fed into the MLLM to produce a response. Model performance is measured by question-answering accuracy. 
For VideoMME, we adopt the temporally diverse regime ($r=1$, $B=m$), partitioning the video into $m$ equal-length segments and selecting one frame per segment to promote uniform temporal coverage.
For LVBench, we use global top-$m$ selection ($r=m$, $B=1$), allocating the frame budget to the highest-scoring candidates across the video without enforcing a per-bin quota.
These two regimes represent the two extremes of the coverage--concentration trade-off controlled by $B$ and $r$ in Eq.~(5). We further cross-evaluate both configurations under the same frame budget in Sec.~\ref{sec:coverage_ablation}.
Following prior work, the evaluations of our method are performed using the \textit{lmms-eval}~\citep{lmms-eval} framework.
\begin{table}[t]
\centering
\small
\setlength{\tabcolsep}{6pt}
\begin{tabular}{lcc}
\toprule
Method & Frames & Overall \\
\midrule
\multicolumn{3}{c}{\textbf{General MLLMs}} \\
\midrule
LLaVA-NeXT-34B~\citep{llavanextvideo}       & 32 & 32.2 \\
InternVL2-40B~\citep{internvl2}             & -- & 39.6 \\
Qwen2-VL-72B~\citep{qwen2vl}                & 48 & 41.3 \\
mPLUG-Owl3-7B~\citep{mplug}                 & 64 & 43.5 \\
InternVL2.5-78B~\citep{internvl2.5}         & 16 & 43.6 \\
Qwen2.5-VL-72B~\citep{qwen25vl}             & -- & 44.0 \\
\midrule
\multicolumn{3}{c}{\textbf{Agentic MLLMs}} \\
\midrule
VideoTree~\citep{videotree}                 & -- & 28.8 \\
VideoAgent~\citep{videoagent}               & -- & 29.3 \\
VCA~\citep{vca}                             & -- & 41.3 \\
\midrule
\multicolumn{3}{c}{\textbf{SFT/RL MLLMs}} \\
\midrule
MovieChat-7B~\citep{moviechat}              & $>$10K    & 22.5 \\
TimeMarker-8B~\citep{timemarker}            & $\leq128$ & 41.3 \\
Video-R1-7B~\citep{videor1}                 & 32        & 37.4 \\
Video-Thinker-7B~\citep{videothinker}       & 32        & 38.4 \\
VideoLLaMA3-7B~\citep{videollama3}          & --        & 45.3 \\
Video-o3~\citep{videoo3}                    & $\leq768$ & 47.6 \\
\midrule
\multicolumn{3}{c}{\textbf{Keyframe Sampling}} \\
\midrule
Qwen2-VL-7B$^\dagger$~\citep{qwen2vl}       & 32 & 39.1 \\
Qwen2-VL-7B + ReTaKe~\citep{Retake}  & $\leq2048$ & 47.8 \\
\rowcolor{ours}
Qwen2-VL-7B + Ours                           & 32 & 49.1 \\
\midrule
Qwen2.5-VL-7B$^\dagger$~\citep{qwen25vl}    & 32 & 37.6 \\
Qwen2.5-VL-7B + FrameThinker~\citep{framethinker}    & 23.9 & 36.6 \\
Qwen2.5-VL-7B + TSPO~\citep{tspo}    & 64 & 46.4 \\
\rowcolor{ours}
Qwen2.5-VL-7B + Ours                      & 32 & 47.7 \\
\midrule
LLaVA-Video-7B$^\dagger$~\citep{llavavideo} & 64 & 41.7 \\
LLaVA-Video-7B + ReTaKe~\citep{Retake}
                                               & $\leq1024$ & 48.5 \\
LLaVA-Video-7B + TSPO~\citep{tspo}    & 64 & 45.3 \\
\rowcolor{ours}
LLaVA-Video-7B + Ours                        & 64 & \textbf{49.4} \\
\bottomrule
\end{tabular}
\caption{Overall accuracy comparison on LVBench~\citep{lvbench}.
Unless otherwise marked, results are taken from the corresponding papers. $^\dagger$ denotes our reproduction of the backbone under our evaluation setting.}
\label{tab:lvbench}
\vspace{-8pt}
\end{table}
\begin{table}[htbp]
\centering
\small
\setlength{\tabcolsep}{5pt}
\begin{tabular}{l c c cc}
\toprule
\multirow{2}{*}{Method} & \multirow{2}{*}{Architecture} & \multirow{2}{*}{Frames} 
& \multicolumn{2}{c}{VideoMME} \\
\cmidrule(lr){4-5}
 &  &  & Long & Average \\
\midrule

Qwen2-VL-7B$^\dagger$ & - & 32 & 48.7 & 58.0 \\
+ AKS~\citep{aks} & BLIP-0.5B & 32 & - & 59.9 \\
+ FOCUS~\citep{focus} & BLIP-0.5B &  32 & - & 59.7 \\
+ Q-Frame~\citep{qframe} & CLIP-0.4B & 44 & 48.3 & 58.3 \\
+ MLLM~\citep{mllm-selector} & MLLM-1.5B & 32 & - & 58.7 \\
\rowcolor{ours} + Ours & CLIP-0.4B  & 32  & \textbf{51.4} & \textbf{60.1}\\
\rowcolor{gray}+ ReTaKe~\citep{Retake} & - & $\leq$ 2048 & 56.2 & 63.9\\
\rowcolor{gray}+ VideoITG ~\citep{videoitg} & MLLM-8B & 56 & 49.2 & 58.8\\
\midrule

Qwen2.5-VL-7B$^\dagger$ & - & 32 & 50.6 & 60.7 \\
+ K-Frames~\citep{kframe} &  MLLM-3B & 32 & - & 62.1 \\
+ AKS~\citep{aks} & CLIP-0.4B & 32 & - & 62.4 \\
+ BOLT~\citep{bolt} & CLIP-0.4B & 32 & - & 62.0 \\
+ ASCS~\citep{kfs} & CLIP-0.4B & 32 & - & 63.1 \\
\rowcolor{ours} + Ours & CLIP-0.4B &  32 & \textbf{55.0} & \textbf{63.6} \\



\bottomrule
\end{tabular}
\caption{Comparison with existing keyframe sampling methods on VideoMME~\citep{videomme}. $\dagger$ denotes our reproduced results.}
\label{tab:videomme_results}
\vspace{-10pt}
\end{table}

\subsection{Comparison with State-of-the-Art Methods}
\textbf{LVBench.} 
LVBench is a long-video understanding benchmark with an average duration of 4101 seconds. 
As shown in Table~\ref{tab:lvbench}, our method achieves competitive performance compared with general MLLMs, agentic MLLMs, SFT/RL-based MLLMs, and several other keyframe sampling approaches. 
Compared with existing keyframe sampling methods~\citep{Retake,tspo,framethinker}, our approach achieves competitive performance under the same MLLM and comparable frame budget.
On fine-grained evaluation, our method shows notable improvements on several subtasks, particularly in key information retrieval, indicating stronger capability in capturing critical temporal cues.


\noindent\textbf{VideoMME.}
As shown in Table~\ref{tab:videomme_results}, our method improves performance on VideoMME under the same MLLM and frame budget. With Qwen2-VL-7B, our approach increases the average accuracy to 60.1 and improves performance on the long-video subset to 51.4, indicating enhanced capability for long-duration video understanding.
With Qwen2.5-VL-7B, our method achieves 63.6 in average accuracy and improves performance on the long-video subset by 4.4 points, demonstrating consistent gains.


\subsection{Generalization across MLLMs}

Beyond the MLLMs used in our main experiments, we further verify that our sampling generalizes to a broader range of
recent and stronger backbones. Specifically, we evaluate three recent MLLMs, i.e., InternVL-3-8B, InternVL-3.5-8B, and Qwen3-VL-8B. As shown in Table~\ref{tab:mllm_backbones}, under a 32-frame budget our method consistently outperforms uniform sampling on LVBench (+8.3, +10.0, and +14.2, respectively) and VideoMME (+1.1, +0.8, and +1.3), confirming that our method brings consistent gains to strong MLLM backbones. Moreover, for the proprietary Gemini-2.5-Flash under a 64-frame budget, our sampling strategy also yields a substantial improvement, 
showing that even strong commercial models benefit from evidential keyframe selection. 

\begin{table}[b]
\centering
\small
\setlength{\tabcolsep}{3pt}
\begin{tabular}{lcccc}
\toprule
Method & Sampling & Frames & LVBench & VideoMME \\
\midrule
\multirow{2}{*}{InternVL-3-8B}
  & Uniform & 32 & 44.0 & 64.9 \\
  & Ours    & 32 & \textbf{52.3} & \textbf{66.0} \\
\midrule
\multirow{2}{*}{InternVL-3.5-8B}
  & Uniform & 32 & 41.0 & 63.7 \\
  & Ours    & 32 & \textbf{51.0} & \textbf{64.5} \\
\midrule
\multirow{2}{*}{Qwen3-VL-8B}
  & Uniform & 32 & 39.3 & 65.6 \\
  & Ours    & 32 & \textbf{53.5} & \textbf{66.9} \\
\midrule
\multirow{2}{*}{Gemini-2.5-Flash}
  & Uniform & 1fps & \textbf{56.7} & - \\
  & Uniform & 64 & 42.6 & - \\
  & Ours    & 64 & 51.8 & - \\
\bottomrule
\end{tabular}
\caption{Generalization across MLLM backbones~\cite{internvl3, internvl3.5, qwen3vl}. Under a
32-frame budget, our evidential keyframe sampling brings
consistent performance improvement to strong MLLM backbones
on LVBench and VideoMME.}
\vspace{-5pt}
\label{tab:mllm_backbones}
\end{table}

\subsection{Generalization across MLLM Scales}

To test the generalization of our method across different
MLLM scales, we evaluate InternVL-3.5~\cite{internvl3.5} at 2B, 8B, and 14B
parameters. As shown in Table~\ref{tab:mllm_scales}, our
sampling method brings consistent performance improvements
over uniform sampling across all model sizes. These results demonstrate that our keyframe sampling strategy is robust to the downstream MLLM scales and generalizes well from small to large models.

\begin{table}[bp]
\centering
\small
\setlength{\tabcolsep}{3pt}
\begin{tabular}{lcccc}
\toprule
Method & Sampling & Frames & LVBench & VideoMME \\
\midrule
\multirow{2}{*}{InternVL-3.5-2B}
  & Uniform & 32 & 37.2 & 57.2 \\
  & Ours    & 32 & \textbf{47.0} & \textbf{58.4} \\
\midrule
\multirow{2}{*}{InternVL-3.5-8B}
  & Uniform & 32 & 41.0 & 63.7 \\
  & Ours    & 32 & \textbf{51.0} & \textbf{64.5} \\
\midrule
\multirow{2}{*}{InternVL-3.5-14B}
  & Uniform & 32 & 43.4 & 64.8 \\
  & Ours    & 32 & \textbf{53.9} & \textbf{65.6} \\
\bottomrule
\end{tabular}
\caption{Ablation study on MLLM scales. We evaluate
InternVL-3.5~\cite{internvl3.5} at 2B, 8B, and 14B parameters. Our method
consistently improves over uniform sampling across all model
scales, with particularly large gains on LVBench.}
\vspace{-5pt}
\label{tab:mllm_scales}
\end{table}

\subsection{Ablation Studies}
\begin{table}[b]
\centering
\small
\setlength{\tabcolsep}{3pt}
\begin{tabular}{ccccc}
\toprule
Method & Sampling & Encoder & VideoMME & LVBench \\
\midrule
\multirow{4}{*}{Qwen2-VL-7B} 
& Uniform & - & 58.0 & 39.1\\
& Ours & CLIP   & \textbf{60.1} & \textbf{49.1} \\
& Ours & SigLIP  & 59.9 & 48.8 \\
& Ours & SigLIP2 & 59.0 & 48.3 \\
\midrule
\multirow{4}{*}{Qwen2.5-VL-7B} 
& Uniform & - & 60.7 & 37.6\\
& Ours & CLIP   & 63.6 & 47.7 \\
& Ours & SigLIP  & \textbf{64.4} & \textbf{48.5} \\
& Ours & SigLIP2 & 61.8 & 43.6 \\
\bottomrule
\end{tabular}
\caption{Ablation study on pretrained vision-language encoders. We evaluate the impact of different encoders within our sampling framework across MLLMs. Our approach consistently surpasses uniform sampling, while the choice of encoder has a moderate impact on the final performance. }
\label{tab:encoder_ablation}
\end{table}
\subsubsection{Impact of Pre-trained Encoders}
We first investigate the influence of different vision-language encoders on overall performance, as shown in Table~\ref{tab:encoder_ablation}. We compare several state-of-the-art encoders, namely CLIP~\citep{clip}, SigLIP~\citep{siglip}, and SigLIP2~\citep{siglip2}.
Our keyframe sampling strategy consistently outperforms uniform sampling across both MLLMs and all encoder choices. The performance, however, varies moderately with the encoder. 
These results indicate that while our method is robust to the choice of encoder, different encoders exhibit varying degrees of alignment with specific MLLMs. Overall, the consistent improvements across all settings demonstrate the generality and effectiveness of our sampling strategy.
\begin{table}
\centering
\begin{tabular}{c c c}
\toprule
\multirow{2}{*}{Frames} & \multicolumn{2}{c}{LVBench} \\
\cmidrule(lr){2-3}
 & Uniform & Ours \\
\midrule
8  & 35.8 & \cellcolor{ours}45.1 ($\uparrow 9.3$) \\
16 & 37.1 & \cellcolor{ours}46.0 ($\uparrow 8.9$) \\
32 & 40.9 & \cellcolor{ours}48.3 ($\uparrow 7.4$) \\
64 & 41.7 & \cellcolor{ours}49.4 ($\uparrow 7.7$) \\
\bottomrule
\end{tabular}
\caption{Ablation study on frame budget. We compare uniform sampling with our method under different numbers of frames on LVBench with LLaVa-Video-7B.}
\vspace{-10pt}
\label{tab:ablation_budget}
\end{table}


\subsubsection{Impact of Keyframe Sampling Budgets}

To investigate how keyframe sampling affects MLLM performance under different computational budgets, we evaluate LLaVA-Video-7B on LVBench with varying numbers of input frames. As shown in Table~\ref{tab:ablation_budget}, our method outperforms uniform sampling across all budgets by a large margin, indicating that our method is able to select more informative frames even under various limited budgets. Moreover, with only 8 sampled frames, our approach already surpasses the performance of uniform sampling with significantly more frames, demonstrating superior efficiency. 

\subsubsection{Impact of Causal Window Size}

The causal window size $w$ controls the extent of local
temporal aggregation in the temporal evidence aggregator~\cref{eq:causal_window}. We study its effect on both
VideoMME and LVBench across two backbones, as shown in
Table~\ref{tab:window_size}. Our method is not very sensitive
to the window size: across $w \in \{2, 4, 8, 16\}$, it
consistently improves over uniform sampling on both
benchmarks, achieving the best performance at
$w = 4$. This observation supports keeping $w$ small relative
to the video length, since an overly large window dissolves
the local neighborhood that evidence segments typically
occupy, diluting the contextualized representation with
temporally distant and largely irrelevant frames. 

\begin{table}[b]
\centering
\small
\setlength{\tabcolsep}{3pt}
\begin{tabular}{lcccc}
\toprule
Method & Sampling & Window & VideoMME & LVBench \\
\midrule
\multirow{5}{*}{Qwen2-VL-7B}
  & Uniform & --  & 58.0 & 39.1 \\
  & Ours    & 2   & 59.1 & 48.9 \\
  & Ours    & 4   & \textbf{60.1} & \textbf{49.1} \\
  & Ours    & 8   & 59.1 & 48.2 \\
  & Ours    & 16  & 59.2 & 48.5 \\
\midrule
\multirow{5}{*}{Qwen2.5-VL-7B}
  & Uniform & --  & 60.7 & 37.6 \\
  & Ours    & 2   & 62.5 & 47.3 \\
  & Ours    & 4   & \textbf{63.6} & \textbf{47.7} \\
  & Ours    & 8   & 62.6 & 46.9 \\
  & Ours    & 16  & 63.3 & 47.4 \\
\bottomrule
\end{tabular}
\caption{Ablation on the causal window size on VideoMME
and LVBench across two backbones. Our method is robust over various window sizes and achieves the best performance
at $w = 4$.}
\vspace{-5pt}
\label{tab:window_size}
\end{table}

\subsubsection{Impact of Selection Rule}
\label{sec:coverage_ablation}
We compare the coverage-oriented ($B=m, r=1$) and concentration-oriented ($B=1, r=m$) regimes using Qwen2.5-VL-7B with 32 frames.
Table~\ref{tab:coverage_ablation} shows that concentration performs better on LVBench, while coverage performs better on VideoMME.
Both outperform uniform sampling; coverage alone improves LVBench accuracy by 7.3 points, indicating that the gains do not depend on concentrated selection. 
Neither rule explicitly penalizes semantic redundancy or guarantees coverage of all complementary evidence.


\begin{table}[t]
\centering
\small
\setlength{\tabcolsep}{5pt}
\begin{tabular}{lcc}
\toprule
Sampling Strategy & LVBench & VideoMME \\
\midrule
Uniform & 37.6 & 60.7 \\
Coverage ($B=m, r=1$) & 44.9 & 63.6 \\
Concentration ($B=1, r=m$) & 47.7 & 61.9 \\
\bottomrule
\end{tabular}
\caption{Ablation on the coverage--concentration trade-off using
Qwen2.5-VL-7B with a fixed 32-frame budget.}
\label{tab:coverage_ablation}
\end{table}

\subsubsection{Impact of Evidential Scoring}
To isolate the benefit of learned evidence scoring from query--frame
semantic relevance, we compare our method with plain CLIP similarity
under the same candidate frames, sampling configuration, and frame
budget. As shown in Table~\ref{tab:clip_ablation}, CLIP-based selection
already improves over uniform sampling, confirming that semantic
relevance provides a strong query-aware signal. Our evidence scorer
further improves accuracy from 47.9 to 49.1 on LVBench and from
59.3 to 60.1 on VideoMME with Qwen2-VL-7B. These consistent gains suggest that
supervision from answer-supporting regions provides complementary
information beyond semantic alignment.
\begin{table}[t]
\centering
\small
\setlength{\tabcolsep}{5pt}
\begin{tabular}{lcc}
\toprule
Sampling Strategy & LVBench & VideoMME \\
\midrule
Uniform & 39.1 & 58.0 \\
CLIP Similarity & 47.9 & 59.3 \\
Ours & \textbf{49.1} & \textbf{60.1}\\
\bottomrule
\end{tabular}
\caption{Comparison with plain CLIP similarity using Qwen2-VL-7B under the
same frame budget and sampling configuration.}
\label{tab:clip_ablation}
\end{table}

\subsection{Efficiency Analysis}
\label{sec:efficiency_analysis}

We evaluate the efficiency--effectiveness trade-off on LVBench using Qwen3-VL-8B-Instruct. Instead of processing thousands of frames, our method selects 300 informative frames and compares against uniform sampling under the same frame budget. We also report the approximately 2K-frame result from the Qwen3-VL technical report~\cite{qwen3vl} as a long-context reference, since this setting exceeded our available GPU memory.

\begin{table}[t]
\centering
\small
\setlength{\tabcolsep}{3pt}
\resizebox{\linewidth}{!}{
\begin{tabular}{c c c c c c}
\toprule
\shortstack[c]{Sampling \\ Strategy} &
\shortstack[c]{Frames \\ Number} &
\shortstack[c]{Accuracy \\ (\%)} &
\shortstack[c]{Sampling \\ Latency} &
\shortstack[c]{Generation \\ Latency} &
\shortstack[c]{Peak \\ Memory} \\
\midrule
Uniform & 300      & 52.6 & --   & 21.0 & $\sim$45 \\
Ours    & 300      & 57.9 & 46.6 & 21.0 & $\sim$45 \\
Uniform & $\sim$2K & 58.0 & --   & --   & -- \\
\bottomrule
\end{tabular}
}
\caption{Efficiency--effectiveness comparison on LVBench using
Qwen3-VL-8B-Instruct. Dashes indicate unavailable or inapplicable
measurements.}
\vspace{-10pt}
\label{tab:qwen3vl_efficiency}
\end{table}

As shown in Table~\ref{tab:qwen3vl_efficiency}, our method improves the 300-frame uniform baseline from 52.6\% to 57.9\%, approaching the reported 58.0\% accuracy with approximately 2K frames. Meanwhile, it uses about 74K visual tokens on average, roughly one-third of the 224K tokens used in the longer-context setting~\cite{qwen3vl}.
The additional sampling latency is 46.6 seconds, mainly due to one-time visual feature extraction at 1~fps. Since these features are query-independent, they can be cached and reused. After caching, query-conditioned scoring takes less than 0.2 seconds per query, making the overhead negligible compared with the 21.0-second MLLM generation latency. Overall, our method shifts computation from repeated long-context MLLM inference to a reusable retrieval stage while maintaining comparable accuracy.

%% file: sec/5_conclusion.tex
\section{Conclusion}
In this work, we proposed a query-conditioned keyframe sampling framework for long-form video understanding from an information-theoretic perspective. 
An idealized conditional mutual information objective and a modular singleton-gain surrogate motivate a tractable frame-wise selection criterion. Guided by this motivation, we learn query-conditioned evidence scores from annotated answer-supporting regions, enabling evidence-based selection beyond conventional semantic-relevance heuristics.
Extensive experiments demonstrate consistent improvements across MLLMs and benchmarks under matched frame budgets, together with a favorable efficiency--effectiveness trade-off. These results highlight the effectiveness of evidential keyframe selection for MLLM-based long-form video understanding.
\newpage